\documentclass{article}

\usepackage{arxiv}

\usepackage[utf8]{inputenc} 
\usepackage[T1]{fontenc}    
\usepackage{hyperref}       
\usepackage{url}            
\usepackage{booktabs}       
\usepackage{amsfonts}       
\usepackage{nicefrac}       
\usepackage{microtype}      
\usepackage{lipsum}
\usepackage{graphicx}

\title{BioNeuralNet: A Graph Neural Network based Multi-Omics Network Data Analysis Tool}
\author{
 Vicente Ramos \\
  Computer Science and Engineering\\
  University of Colorado Denver\\
  \texttt{vicente.ramos@ucdenver.edu} \\
  \And
 Sundous Hussein \\
  Computer Science and Engineering\\
  University of Colorado Denver\\
  \texttt{sundous.hussein@ucdenver.edu} \\
  \And
 Mohamed Abdel\mbox{-}Hafiz \\
  Computer Science and Engineering\\
  University of Colorado Denver\\
  \texttt{mohamed.abdel-hafiz@ucdenver.edu} \\
  \And
 Arunangshu Sarkar \\
  Biostatistics and Bioinformatics\\
  University of Colorado Anschutz Medical Campus\\
  \texttt{arunangshu.sarkar@cuanschutz.edu} \\
  \And
 Weixuan Liu \\
  Biostatistics and Bioinformatics\\
  University of Colorado Anschutz Medical Campus\\
  \texttt{weixuan.liu@cuanschutz.edu} \\
  \And
 Katerina J.\ Kechris \\
  Biostatistics and Bioinformatics\\
  University of Colorado Anschutz Medical Campus\\
  \texttt{katerina.kechris@cuanschutz.edu} \\
  \And
 Russell P.\ Bowler \\
  Genomic Medicine Institute\\
  Cleveland Clinic Main Campus\\
  \texttt{bowlerr@ccf.org} \\
  \And
 Leslie Lange \\
  Biomedical Informatics and Personalized Medicine\\
  University of Colorado Anschutz Medical Campus\\
  \texttt{leslie.lange@cuanschutz.edu} \\
  \And
 Farnoush Banaei-Kashani \\
  Computer Science and Engineering\\
  University of Colorado Denver\\
  \texttt{farnoush.banaei-kashani@ucdenver.edu} \\
}

\begin{document}
\maketitle
\begin{abstract}
Multi-omics data offer unprecedented insights into complex biological systems, yet their high dimensionality, sparsity, and intricate interactions pose significant analytical challenges. Network-based approaches have advanced multi-omics research by effectively capturing biologically relevant relationships among molecular entities. While these methods are powerful for representing molecular interactions, there remains a need for tools specifically designed to effectively utilize these network representations across diverse downstream analyses. To fulfill this need, we introduce \textit{BioNeuralNet}, a flexible and modular Python framework tailored for end-to-end network-based multi-omics data analysis. \textit{BioNeuralNet} leverages Graph Neural Networks (GNNs) to learn biologically meaningful low-dimensional representations from multi-omics networks, converting these complex molecular networks into versatile embeddings. \textit{BioNeuralNet} supports all major stages of multi-omics network analysis, including several network construction techniques, generation of low-dimensional representations, and a broad range of downstream analytical tasks. Its extensive utilities, including diverse GNN architectures, and compatibility with established Python packages (e.g., scikit-learn, PyTorch, NetworkX), enhance usability and facilitate quick adoption. \textit{BioNeuralNet} is an open-source, user-friendly, and extensively documented framework designed to support flexible and reproducible multi-omics network analysis in precision medicine.\\
\textbf{Availability and implementation:} \textit{BioNeuralNet} is available via The Python Package Index (PyPI). Source code, documentation, tutorials, and workflows are hosted at \url{https://bioneuralnet.readthedocs.io}.

\end{abstract}

\section{Introduction}
Recent advancements in multi-omics technologies have facilitated simultaneous profiling of genomics, transcriptomics, proteomics, and metabolomics, significantly deepening our understanding of complex biological systems. Yet, effectively extracting actionable insights from these high-dimensional, sparse datasets remains challenging due to intricate molecular interactions and inherent variability.

Network-based approaches, such as Weighted Gene Co-expression Network Analysis (WGCNA) \cite{Langfelder2008} and Sparse Multiple Canonical Correlation Network (SmCCNet) \cite{Liu2024}, have been instrumental in identifying biological modules and key molecular interactions. While these methods effectively capture molecular relationships, they are not always optimized for translating those networks into actionable data representations for downstream tasks. There is a growing need for tools that can build upon these network representations, unlocking their potential for flexible, scalable, and diverse analytical applications.

Building upon the strengths of multi-omics networks, we introduce BioNeuralNet, a flexible, modular Python framework leveraging Graph Neural Networks (GNNs) to transform multi-omics networks into biologically meaningful low-dimensional embeddings. These embeddings distill complex, nonlinear molecular relationships into compact vectorized representations, enabling effective and efficient implementation of a wide range of downstream tasks such as disease prediction, biomarker discovery, and subject-level profiling with improved accuracy and scalability.

BioNeuralNet also supports integration of phenotype and clinical data, further enhancing the biological and clinical relevance of the generated embeddings. Its modular design ensures adaptability to diverse research scenarios, compatibility with existing Python packages (e.g., scikit-learn, NetworkX, Matplotlib), and ease of use through extensive documentation and illustrative examples.

In the subsequent sections, we present the BioNeuralNet architecture, perform a comparative evaluation of its effectiveness against existing leading multi-omics methods, and demonstrate practical applications enabled by BioNeuralNet-generated representations.

\begin{figure}
    \centering
    \includegraphics[width=\textwidth]{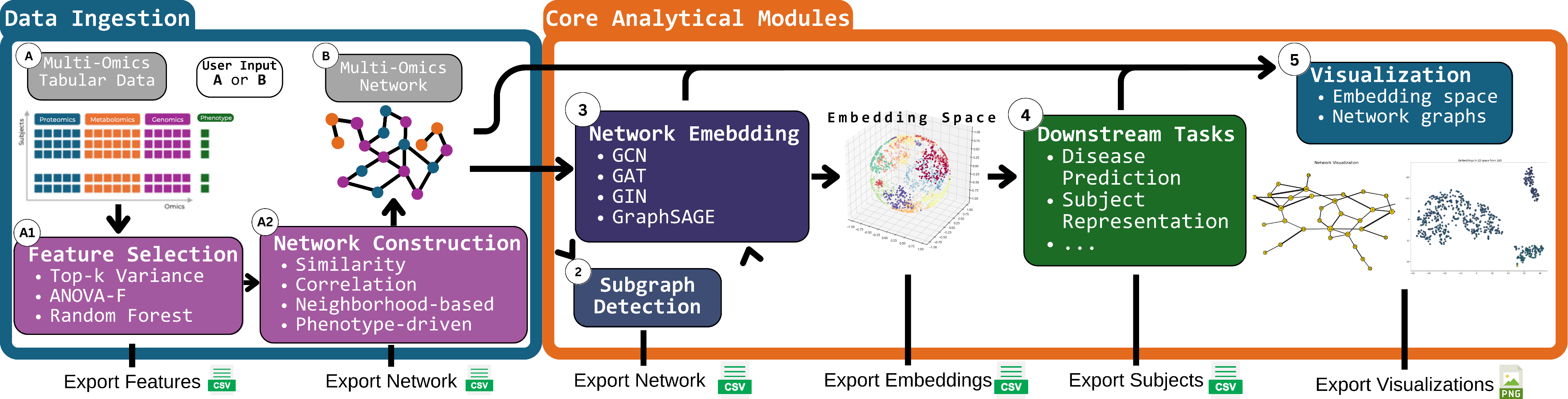}
    \caption{\textbf{BioNeuralNet workflow overview.} Users may begin with either multi-omics tabular data (A) or an existing multi-omics network (B). Tabular data can undergo feature selection (A1) and network construction (A2) to construct a network representation. The core analytical modules (orange border) include optional subgraph detection (2), network embedding using various GNN architectures (3), downstream tasks such as disease prediction (4), and visualizations (5). At each stage, intermediate and final outputs can be exported in CSV or PNG format. The embedding space image is adapted from \cite{McInnes2018}.}
    \label{fig:pipeline}
\end{figure}

\section{Related Work}

Existing approaches to multi-omics analysis can be broadly categorized into two groups: non-network-based statistical methods, and task-specific network-based models.

Traditional statistical methods typically represent multi-omics data as high-dimensional tabular matrices. Methods such as Multi-Omics Factor Analysis (MOFA)  \cite{Argelaguet2018} extract shared latent structures across omics modalities in an unsupervised manner. While these approaches are effective at capturing global variation, they generally overlook the relationships and interactions between biomolecular entities, resulting in limited biological interpretability and downstream analytical flexibility.

In contrast, several recent methods employ network-based deep learning to explicitly model multi-omics as biological networks for phenotype prediction. Representative examples include MOGONET \cite{Wang2021} and SUPREME  \cite{Kesimoglu2023}, which leverage Graph Convolutional Networks (GCNs) to build supervised classifiers. While these methods demonstrate strong predictive performance for specific tasks such as cancer subtype classification, they are often designed as task-specific pipelines, their architectures are typically coupled to particular GNN variants, and they fail to provide the extensibility and modularity needed for broader exploratory analysis, integration of diverse network types, or adaptation to new tasks.

BioNeuralNet differs fundamentally in that it is not a task-specific model, but rather a general, flexible, user-oriented tool for analysis of multi-omics networks. It enables researchers to integrate external biological knowledge, select or implement different GNN architectures, and apply the system across a wide range of supervised and unsupervised analyses. To our knowledge, no existing open-source tool provides this level of adaptability, reproducibility, and ease of use for multi-omics network representation learning and analysis.

\section{System Overview}

BioNeuralNet is a modular framework that streamlines every stage of network-based multi-omics analysis, from initial data import and network construction to representation learning and downstream analysis \ref{fig:pipeline}. The system emphasizes flexibility, interoperability, and ease of use, allowing researchers to tailor each step to their experimental context.

\subsection{Data Ingestion}

BioNeuralNet users can provide their own precomputed multi-omics network, or alternatively upload tabular omics matrices in DataFrame or CSV format (along with optional phenotype or clinical annotations) and use BioNeuralNet's built-in network construction routines to construct their network defined as follows: \verb+G = (V, E)+, where:
\begin{itemize}
    \item $V$ is the set of nodes representing one or more omics modalities;
    \item $E \subseteq V \times V$ is the set of edges describing biological relationships, such as co-expression or similarity;
    \item Each node $v \in V$ may carry a feature vector representing molecular measurements or attributes, while each edge may have weights or types indicating the strength or category of association.
\end{itemize}

For users starting from raw matrices, BioNeuralNet provides a variety of network-construction strategies, including:

\begin{itemize}
    \item \textit{Similarity Networks} (e.g., cosine or Euclidean similarity, with optional pruning and cutoffs);
    \item \textit{Correlation Networks} (e.g., Pearson or Spearman correlations, or WGCNA-style soft thresholding);
    \item \textit{Neighborhood-based Networks} (e.g., k-nearest neighbor or shared-nearest neighbor Networks);
    \item \textit{Phenotype-driven Networks} (e.g., SmCCNet for supervised multi-omics network inference \cite{Liu2024}.
\end{itemize}

A complete list of supported network construction methods and their parameters is provided in the documentation of BioNeuralNet. The flexibility in data ingestion allows investigators to benchmark different network topologies and select the approach that best captures the biological relationships within their system of interest.

Prior to network construction, BioNeuralNet provides data cleaning and dimensionality reduction utilities to support both supervised and unsupervised workflows. Supervised options include random forest feature importance, ANOVA-F tests (for categorical or continuous outcomes), and correlation-based ranking (supervised correlation against target outcomes). Unsupervised tools include retention of high-variance feature. Additional network-based pruning utilities are available to refine node features or edge connectivity. Further details and examples are provided in the documentation of BioNeuralNet.

\subsection{Core Analytical Modules}

\subsubsection{Subgraph Detection}
BioNeuralNet supports both supervised and unsupervised community detection methods to identify biologically meaningful subgraphs within omics networks \cite{Abdel-Hafiz2022}. Its core clustering module implements multi-omics extensions of the Louvain algorithm, personalized PageRank, and hybrid approaches, allowing flexible detection of dense feature modules associated with phenotypes or experimental conditions. These methods leverage both network connectivity and sample-level data to reveal coherent groups of genes, proteins, or metabolites that may correspond to pathways, functional modules, or disease mechanisms. Output subgraphs can be further analyzed for enrichment, linked to clinical outcomes, or used to reduce dimensionality prior to downstream machine learning tasks. Full details of available clustering strategies and their customization are provided in the documentation of BioNeuralNet.

\subsubsection{Network Embedding}
BioNeuralNet supports a range of Graph Neural Network (GNN) architectures for embedding generation, including Graph Convolutional Networks (GCN), Graph Attention Networks (GAT), GraphSAGE, and Graph Isomorphism Networks (GIN) \cite{Scarselli2009}. Each model is suited to different types of biological networks and analysis goals. For example, GCNs are effective in the case of uniformly connected graphs, GATs use attention mechanisms to highlight key biological relationships, GraphSAGE is designed for large or dynamic datasets, and GIN is sensitive to subtle feature variations. Users can select from these integrated models within BioNeuralNet or extend the framework with additional architectures that follow the PyTorch Geometric interface \cite{paszke2019pytorch}. This modular approach allows embeddings to capture both network structure and biological context, supporting a wide range of downstream analyses. 

\subsubsection{Downstream Tasks}
BioNeuralNet enables a broad range of downstream analyses using the generated rich network embeddings, with several key workflows implemented and additional applications easily supported:

\begin{itemize}
    \item Disease prediction:
    BioNeuralNet integrates Disease Prediction using Multi-Omics Networks (DPMON) \cite{Hussein2024} for end-to-end supervised disease classification through network embeddings. DPMON combines adjacency networks, multi-omics data, and optional clinical covariates, with support for hyperparameter tuning across GCN, GAT, GraphSAGE, and GIN architectures.

    \item Enhanced subject representation:
    BioNeuralNet produces low-dimensional embeddings for each subject leveraging the network-driven representations of the omics instead of the raw omics readings for each sample \cite{Hussein2022}. The network-driven subject representation enables enhanced subject subtyping, clustering, biomarker discovery, subject stratification, and visualization.
    \item Other user-defined tasks:
    BioNeuralNet embeddings can be used for implementation of a variety of additional downstream tasks, such as omics-omics interaction prediction, functional annotation, pathway discovery, subject similarity analysis, progression trajectory modeling, and visualization of complex
    multi-omics relationships (e.g., using UMAP \cite{McInnes2018}) to embed the input multi-omics networks. This extensibility highlights the framework's flexibility for integrative multi-omics research and supports the rapid development of new analytical workflows by BioNeuralNet users.
\end{itemize}

BioNeuralNet is designed for extensibility, allowing users to implement custom downstream tasks by leveraging its modular API. Additional modules and workflow examples are described in the BioNeuralNet documentation, in particular in the Availability section.

\section{Demonstrative Workflow}

To illustrate the practical application and flexibility of BioNeuralNet, we present an example workflow using the TCGA-BRCA breast cancer dataset \cite{cancer}. This case study demonstrates how BioNeuralNet supports multi-omics data harmonization, feature engineering, network construction, and disease prediction. All code and workflow scripts for this demonstrative workflow are available in the BioNeuralNet documentation. Initially, the dataset contained DNA methylation (20,107 features × 885 samples), mRNA (18,321 features × 1,212 samples), miRNA (503 features x 1,189 samples), clinical (101 variables × 1,098 samples), and PAM50 subtype labels. Rigorous quality control included barcode standardization, aggregation of multiple aliquots per subject (by averaging), and removal of incomplete cases. This process aligned all modalities, resulting in a unified cohort of 769 subjects with comprehensive and high-quality data across all molecular and clinical features: DNA methylation (769 × 20,107), mRNA (769 × 20,531), miRNA (769 × 503), clinical (769 × 119), and PAM50 labels. Below we described the steps we take to analyze the prepared data using BioNeuralNet:

\paragraph{Step 1: Feature Selection}
To address the curse of dimensionality, three feature selection approaches were applied and compared on mRNA and DNA methylation matrices: variance thresholding, ANOVA F-test, and random forest importance. For each modality, the top 6,000 features were retained. All miRNA features ($n=503$) were kept, and clinical variables were reduced to the ten most predictive features using random forest feature importance analysis to incorporate clinical covariates in the predictive model. The overlap among selection methods  highlights their complementary capabilities and robustness (see Table \ref{tab:preprocessing}).

\paragraph{Step 2: Network Construction}
A $k$-nearest neighbor ($k=15$) cosine similarity network is constructed over selected features, capturing functional relationships between genes, DNA methylation, and miRNAs. BioNeuralNet's modular interface supports rapid iteration on network construction strategies and easy integration of new omics data types.

\paragraph{Step 3: GNN-Based Disease Prediction}
BioNeuralNet enables end-to-end network-based phenotype prediction by integrating omics data, clinical variables, and network structure within a single, user-friendly function. This is implemented by \texttt{DPMON} module within BioNeuralNet, which follows a four-step workflow: (1) a Graph Neural Network (GNN) extracts informative feature embeddings from multi-omics networks, capturing both local and global relationships among features; (2) dimensionality reduction is applied to these embeddings using averaging, maximum, or autoencoder techniques to reduce computational complexity while preserving meaningful information; (3) the reduced embeddings are integrated with the original multi-omics data through concatenation or feature weighting, enhancing each sample's representation; and (4) a feed-forward neural network utilizes the integrated feature set to predict phenotypes, with architecture and training optimized jointly with the GNN for improved accuracy. All steps are configurable, allowing users to tailor the analysis while benefiting from BioNeuralNet streamlined workflow.

The results of the aforementioned analysis are shown in (Table \ref{tab:results}). As demonstrated, BioNeuralNet significantly
outperforms a set of representative models from the literature.

\begin{table}[!t]
\caption{Feature Overlap Across Methods\label{tab:preprocessing}}%
\begin{tabular*}{\columnwidth}{@{\extracolsep\fill}lll@{\extracolsep\fill}}
\toprule
\textbf{Combination} & \textbf{mRNA} & \textbf{DNA methylation} \\
\midrule
ANOVA F $\cap$ Variance & 2,340 & 2,091 \\
Random Forest $\cap$ Variance & 2,218 & 1,871 \\
ANOVA F $\cap$ Random Forest & 2,546 & 2,201 \\
All Three Methods & 1,134 & 815 \\
\bottomrule
\end{tabular*}
\end{table}

\begin{table}[!t]
\caption{Results on TCGA-BRCA subtype classification. Random Forest and MOGONET values were taken from the MOGONET study, whereas MLP and SUPREME values were taken from the SUPREME study. Scores reflect each method best-reported performance. F1-weighted and F1-macro, accounting for class imbalance and equal class weighting respectively. \label{tab:results}}%
\begin{tabular*}{\columnwidth}{@{\extracolsep\fill}llll@{\extracolsep\fill}}
\toprule
\textbf{Method} & \textbf{Accuracy} & \textbf{F1 Weighted} & \textbf{F1 Macro} \\
\midrule
RandomForest & $0.789 \pm 0.018$ & $0.786 \pm 0.019$ & $0.722 \pm 0.020$ \\
MOGONET & $0.829 \pm 0.018$ & $0.825 \pm 0.016$ & $0.774 \pm 0.017$ \\
MLP & $0.820 \pm 0.030$ & $0.820 \pm 0.030$ & $0.820 \pm 0.030$ \\
SUPREME & $0.840 \pm 0.010$ & $0.830 \pm 0.010$ & $0.750 \pm 0.010$ \\
BioNeuralNet & $0.951 \pm 0.039$ & $0.937 \pm 0.053$ & $0.841 \pm 0.163$ \\
\bottomrule
\end{tabular*}
\end{table}

\section{Conclusion}
BioNeuralNet is a flexible and modular Python framework designed to address the key challenges of multi-omics data analysis. By integrating advanced Graph Neural Network architectures with robust feature selection and user-friendly interfaces, BioNeuralNet enables researchers to extract actionable biological insights from complex, high-dimensional datasets. Its open-source design, comprehensive documentation, and demonstrated performance across diverse analytical tasks establish BioNeuralNet as a leading platform for reproducible, scalable, and accessible multi-omics network analysis in precision medicine and systems biology.

\section*{Acknowledgments}
{Research reported in this work was supported by the National Heart, Lung, and Blood Institute of the National Institutes of Health under award number R01HL152735.}

\bibliographystyle{unsrt}  



\end{document}